\documentclass[runningheads]{llncs}
\usepackage[T1]{fontenc}
\usepackage{graphicx}
\usepackage{booktabs}
\usepackage{amsmath}
\usepackage{amssymb}
\usepackage{multirow}
\usepackage{subcaption}
\usepackage{xcolor}
\usepackage{cite}
\usepackage{orcidlink}
\usepackage[utf8]{inputenc}
\usepackage{comment}
\usepackage{url}

\begin{document}

\title{Now You Have My Healthy Attention:\\ A U-DiT for Brain-MRI Inpainting}
\titlerunning{Now You Have My Healthy Attention}

\author{Danilo Danese\orcidlink{0009-0000-5203-1229} \and
Angela Lombardi\orcidlink{0000-0003-1815-9522} \and
Tommaso Di Noia\orcidlink{0000-0002-0939-5462}}
\authorrunning{D. Danese et al.}
\institute{Politecnico di Bari, Bari, Italy\\
\email{\{danilo.danese, angela.lombardi, tommaso.dinoia\}@poliba.it}}

\maketitle

\begin{abstract}
The ASNR-MICCAI BraTS Local Synthesis (Inpainting) task asks for the anatomically plausible completion of healthy brain tissue within a masked region of a T1-weighted MRI, providing a tumor-free anatomical reference for downstream analysis. As the task is scored by distortion metrics (SSIM, PSNR, MSE), we build a deterministic regression model and focus on giving it inductive biases tailored to inpainting. Our network follows the U-DiT principle of performing self-attention on a downsampled token grid: a volumetric encoder-decoder imports long-range context through a downsampled global self-attention block with three-dimensional rotary position embeddings, while convolutions and skip connections preserve high-frequency detail. Two ideas drive our results. First, we constrain the attention so that occluded ("void") tokens attend only to known-healthy tokens of the same volume, with a learned bias toward each query's contralateral homologue, forcing the completion to be inferred from observed anatomy rather than from other unknown regions. Second, we add a contralateral-symmetry input that supplies the mirrored healthy hemisphere as a patient-specific prior; since the brain is approximately bilaterally symmetric and lesions are typically unilateral, this prior improves the distortion metrics at matched structural similarity. On the official BraTS-2026 validation leaderboard our submission reaches a mean healthy-region SSIM of $0.864$, PSNR of $24.7$\,dB and MSE of $4.6{\times}10^{-3}$ over $219$ cases. We further analyse the residual smoothness inherent to distortion-optimal regression and discuss its implications for anatomical realism.
\end{abstract}

\section{Introduction}\label{intro}

Quantitative analysis of brain tumours from magnetic resonance imaging (MRI) supports diagnosis, surgical planning and treatment monitoring. A recurring obstacle is that a patient-specific healthy reference essentially never exists: the first scan is acquired only after symptom onset, so the anatomical baseline is already compromised. This absence injects a "pathology bias" into registration, atlas construction and segmentation pipelines. Synthesising the missing healthy anatomy in the lesioned region reconstructs a plausible healthy anatomical reference and enables downstream analyses that would otherwise be confounded by the lesion. The ASNR-MICCAI BraTS \emph{Local Synthesis of Healthy Brain Tissue via Inpainting} task formalises this problem~\cite{kofler2023inpainting}: given a T1-weighted volume with a masked region, a model must inpaint plausible healthy tissue, evaluated over the healthy portion of the mask by the Structural Similarity Index (SSIM)~\cite{wang2004ssim}, peak signal-to-noise ratio (PSNR) and mean-squared error (MSE).

Prior BraTS inpainting methods fall into two main families. Regression models trained with pixel-wise and SSIM losses~\cite{zhang2023synthesis,zhang2024unet,zhang2025robust} predict a single completion and achieve strong performance on distortion-based metrics. A recurring ingredient is aggressive random-mask augmentation, which exposes the network to diverse mask shapes and locations during training~\cite{zhang2025robust}. In contrast, generative approaches, particularly diffusion models~\cite{durrer2024denoising,ferreira2024wdm,durrer2025fastwdm3d}, sample from the conditional distribution and generate more realistic textures, but typically sacrifice distortion metrics in favour of perceptual quality. We therefore focus on the regression setting and investigate which inpainting-specific architectural priors, rather than generic image-to-image backbones, most effectively improve reconstruction quality. Recent diffusion transformers replace convolutional feature extraction with self-attention over patch tokens~\cite{peebles2023dit} and have shown promising results for 3D brain MRI synthesis, including wavelet-domain flow-matching frameworks~\cite{danese2026wavedit}. Among them, U-DiT improves computational efficiency by applying attention on a downsampled token grid while preserving most of its representational power~\cite{tian2024udit}. Finally, we leverage the approximate bilateral symmetry of the brain, a well-established prior for lesion analysis~\cite{ma2024symmetry}, both by providing the contralateral mirrored hemisphere as an additional input channel and by applying mirror test-time augmentation during inference~\cite{wang2019tta}.

Because these metrics are optimised by the conditional mean of plausible completions, we adopt a deterministic regression model and focus on the inductive biases most relevant to image inpainting. We build on the U-DiT idea of performing self-attention on a downsampled token grid~\cite{tian2024udit}, which brings global context to a volumetric network at a fraction of the cost of full-resolution attention, and ask what additional structure the inpainting task itself demands. We identify two key inductive biases for this task: reconstruction should rely only on observed healthy tissue, and the contralateral hemisphere provides a strong patient-specific anatomical prior.

Our contributions are:
\begin{itemize}
  \item A \textbf{volumetric network based on the U-DiT architecture}: a downsampled global self-attention block with three-dimensional rotary position embeddings~\cite{su2024roformer,tian2024udit} captures long-range contextual dependencies, while convolutions and skip connections carry the local anatomical detail.
  \item A \textbf{non-local "healthy-only" attention with contralateral weighting} that removes occluded tokens from the set of attendable keys, so a query in the void attends exclusively to known-healthy tokens of the same volume, among which a learned, zero-initialised bias peaked at the query's mirror position decides which to read first. This makes the reconstruction draw on observed anatomy rather than on other, unknown regions.
  \item A \textbf{contralateral-symmetry input} that feeds the network the left-right mirror of the voided volume about the estimated mid-sagittal plane, together with a validity channel that discounts mirror voxels that fall outside the brain or inside the void. The contralateral healthy hemisphere provides a strong patient-specific anatomical prior for reconstructing the missing tissue.
\end{itemize}

Code is available at \url{https://github.com/sisinflab/WaveUDiT}.

\section{Methods}\label{methods}

\subsection{Dataset and pre-processing}
We use the BraTS-Local-Inpainting data~\cite{kofler2023inpainting},   derived from the BraTS-GLI T1-weighted collection~\cite{baid2021rsna,bakas2017advancing,menze2015brats,bakas2017tcgagbm,bakas2017tcgalgg} ($1{,}251$ training cases). Each case provides the ground-truth T1n, a voided image with a masked region removed and the binary mask; scoring is restricted to the healthy sub-region of the mask. We normalise intensities by the per-volume maximum to $[0,1]$, rescale to $[-1,1]$, and crop a $144{\times}208{\times}208$ window around the void for training. At inference the known tissue outside the mask is copied back verbatim, so only the void is synthesised.

\begin{figure}[tb]
  \centering
  \includegraphics[width=1\textwidth]{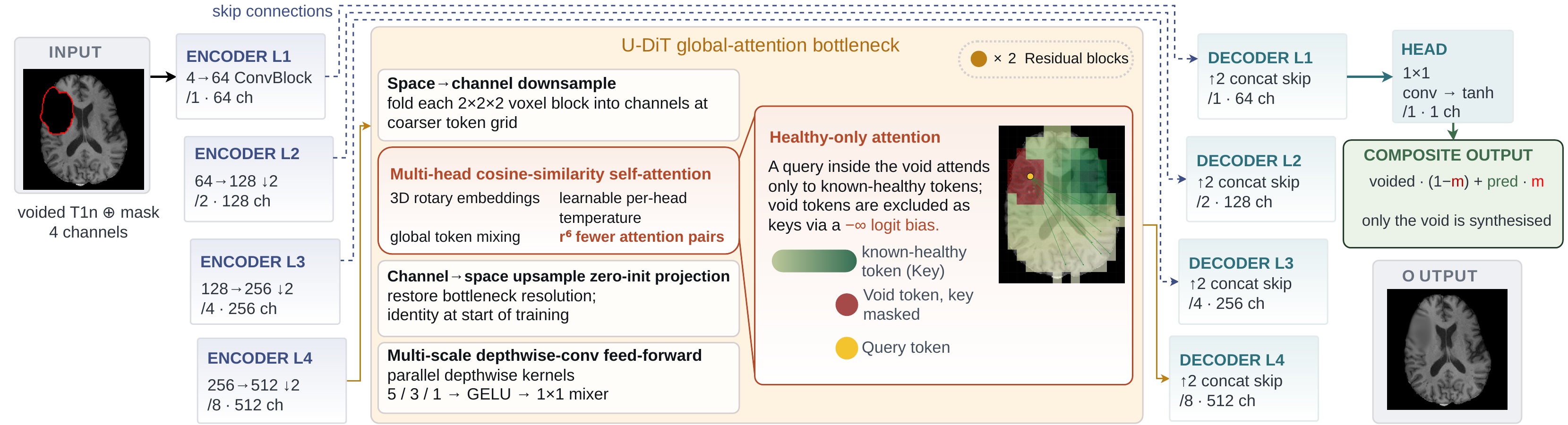}
  \caption{The U-DiT-style volumetric network. A convolutional encoder--decoder with skip connections carries local detail; the bottleneck holds a single downsampled global self-attention block (space-to-channel $\downarrow r$, cosine-similarity attention with 3D RoPE, channel-to-space $\uparrow r$).}
  \label{fig:arch}
\end{figure}

\subsection{Network architecture}
The network is a 3D convolutional encoder-decoder (base width $64$, four resolution levels, GroupNorm/SiLU) that injects global context through a single U-DiT block~\cite{tian2024udit} at the bottleneck (Fig.~\ref{fig:arch}). The U-DiT principle is to run self-attention not at full resolution but on a {downsampled} token grid, which yields long-range token mixing at a small fraction of the cost of full-resolution attention while the surrounding convolutions and skip connections preserve high-frequency detail. Concretely, we apply a lossless space-to-channel reshape, the 3D analogue of pixel-unshuffle, that stacks each $r{\times}r{\times}r$ block of neighbouring voxels ($r{=}2$) into the channel dimension: a $C$-channel grid of size $D{\times}H{\times}W$ becomes an $r^3C$-channel grid of size $\tfrac{D}{r}{\times}\tfrac{H}{r}{\times}\tfrac{W}{r}$ with no loss of information. The token grid, therefore, holds $r^3{=}8\times$ fewer positions, and since self-attention compares every pair of tokens its cost drops by $r^6{=}64\times$. A $1{\times}1$ convolution then maps the widened channels to the token dimension, multi-head cosine-similarity attention with a learnable per-head temperature and 3D rotary position embeddings~\cite{su2024roformer} mixes the tokens globally, and a channel-to-space reshape (the inverse pixel-shuffle) restores the resolution. The attention output projection is zero-initialised, so the block is an identity at initialisation and the network begins as its convolutional counterpart. Every attention operation in the model is this single coarse-scale block, which supplies exactly the non-local context a convolutional receptive field lacks. The network predicts the full volume with a $\tanh$ head, composited as
\begin{equation}
  \hat{x} \;=\; x_v \odot (1-m) \;+\; r_\theta(x_v, m)\odot m,
\end{equation}
with $x_v$ the voided input, $m$ the binary mask (one inside the void, zero on known tissue), and $r_\theta$ the network, so known tissue is preserved exactly.

\paragraph{Design rationale: why not a full transformer.}
Our architecture builds on a wavelet-domain diffusion transformer for 3D brain MRI synthesis~\cite{danese2026wavedit}, based on the hierarchical hourglass transformer~\cite{crowson2024hdit}. Rather than adopting the full transformer backbone, we retain only its downsampled-attention primitive and integrate it into a convolutional hierarchy, which we found better suited to brain MRI inpainting.
In preliminary experiments, a full-transformer backbone plateaued around $0.58$ SSIM on our internal split, well below the convolutional model, at substantially higher compute (Section~\ref{results}). This design is consistent with the nature of the task. The non-local information required for inpainting, such as global anatomy and contralateral context, is predominantly low-frequency and can therefore be captured by a single global-attention block at the bottleneck. In contrast, the convolutional hierarchy preserves fine anatomical detail and provides a stronger inductive bias than a full transformer when training on the available cases with a distortion-based objective.

\paragraph{Tokenizer.}
The linear bottleneck patchify is augmented with a residual overlapping-convolution branch: a band-grouped convolution for the token merge and a depthwise convolution before a pixel-shuffle~\cite{shi2016espcn} for the split, giving each token cross-boundary spatial context. The branch output projection is zero-initialised, so it is inactive at initialisation and preserves the identity start.

\subsection{Non-local healthy-only attention with contralateral weighting}
\begin{figure}[tb]
  \centering
  \includegraphics[width=1\textwidth]{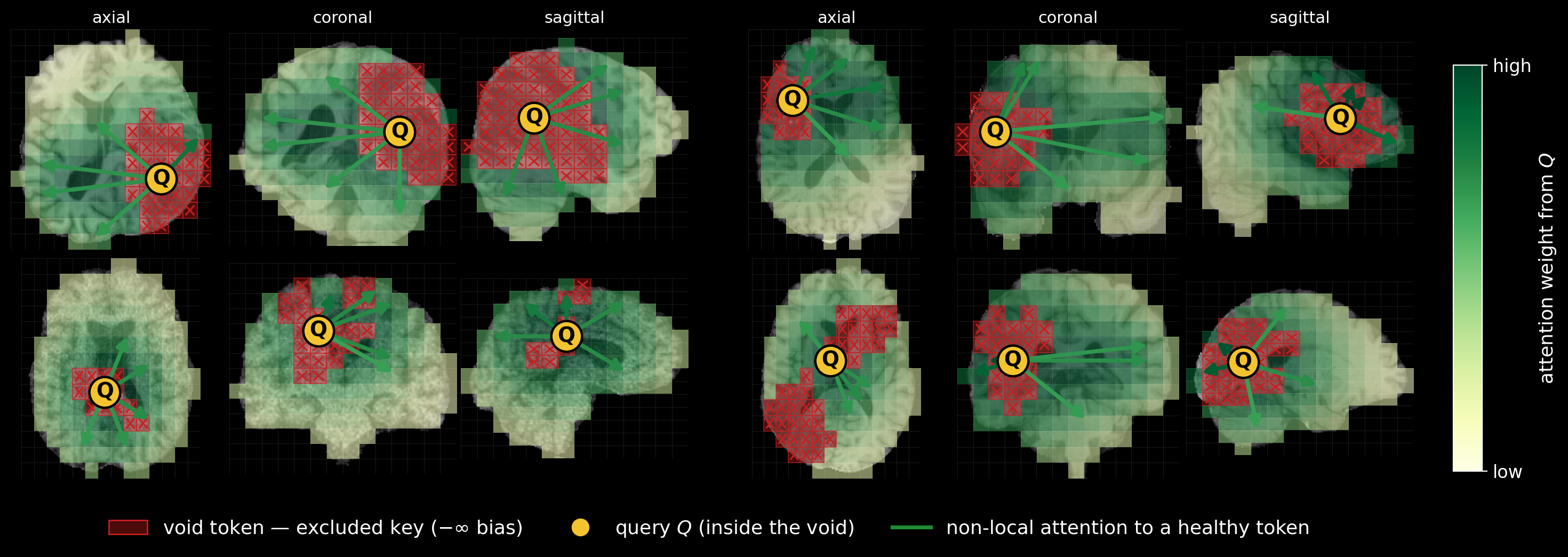}
  \caption{Healthy-only attention with contralateral weighting, shown for four validation cases (rows), each in three anatomical views (columns). Void tokens (red $\times$) receive a $-\infty$ key bias and are therefore excluded as keys, so a query $Q$ inside the void (gold) reads only known-healthy tokens (green, shade $=$ attention weight) and can reach them non-locally across the whole volume. In the axial and coronal views, where the left-right axis lies in the plane, the learned bias visibly concentrates the weight on the homologous region of the opposite hemisphere; the sagittal view shows the healthy-only weighting alone, since there the homologue falls outside the displayed slice.}
  \label{fig:attn}
\end{figure}

Left unconstrained, the bottleneck attention lets a query inside the void pool information from other void tokens, content that is itself unknown and being predicted, which turns reconstruction into a fixed-point over missing values. We remove this coupling, and at the same time tell each query which of the remaining tokens to prefer, through a single additive term on the attention logits. For a query token $i$ and key token $j$,
\begin{equation}
\ell_{ij}=s\,\frac{q_i^{\top}k_j}{\lVert q_i\rVert\,\lVert k_j\rVert}+b_j
      +\lambda_h\,\bar v_j\,\exp\!\Big(-\tfrac{\lVert p_j-\mu(p_i)\rVert^{2}}{2\sigma^{2}}\Big),
\end{equation}
with a learnable per-head temperature $s$, normalised into attention weights $\alpha_{ij}=\mathrm{softmax}_j(\ell_{ij})$. The bias $b_j$ is read off the mask $m$ average-pooled onto the coarse token grid, each token aggregating the $r^3$ voxels of its own block: writing $\bar m_j$ for the resulting per-token void fraction, $b_j=-\infty$ where $\bar m_j>0.5$ and $b_j=0$ elsewhere, so every predominantly-void key receives zero weight and each query, void or observed, attends only to known-healthy tokens of the same volume, reaching non-locally across both hemispheres (Fig.~\ref{fig:attn}). A guard disables the bias for the degenerate all-void sample to avoid an undefined $\mathrm{softmax}$. This part is parameter-free and, as Section~\ref{results} shows, the most effective single change we make.

The last term then decides {which} admissible key to read first, since the healthy tissue that best predicts a void voxel is usually its own contralateral homologue~\cite{ma2024symmetry}: $p_i$ is the grid position of token $i$, $\mu(p_i)$ its reflection about the estimated mid-sagittal plane, and the per-head strength $\lambda_h$ and width $\sigma$ are learned. Three properties make it safe to add to an already-trained network. The plane is the same per-volume estimate used for the contralateral input channel rather than the centre of the token grid, so the preference follows the anatomy and not the field of view. The gate $\bar v_j$ is the pooled mirror-validity map, which suppresses the term exactly where the mirror falls outside the brain or inside the void, as happens for midline-crossing or bilateral lesions.

\subsection{Contralateral-symmetry input}

\begin{figure}[tb]
  \centering
  \includegraphics[width=0.6\textwidth]{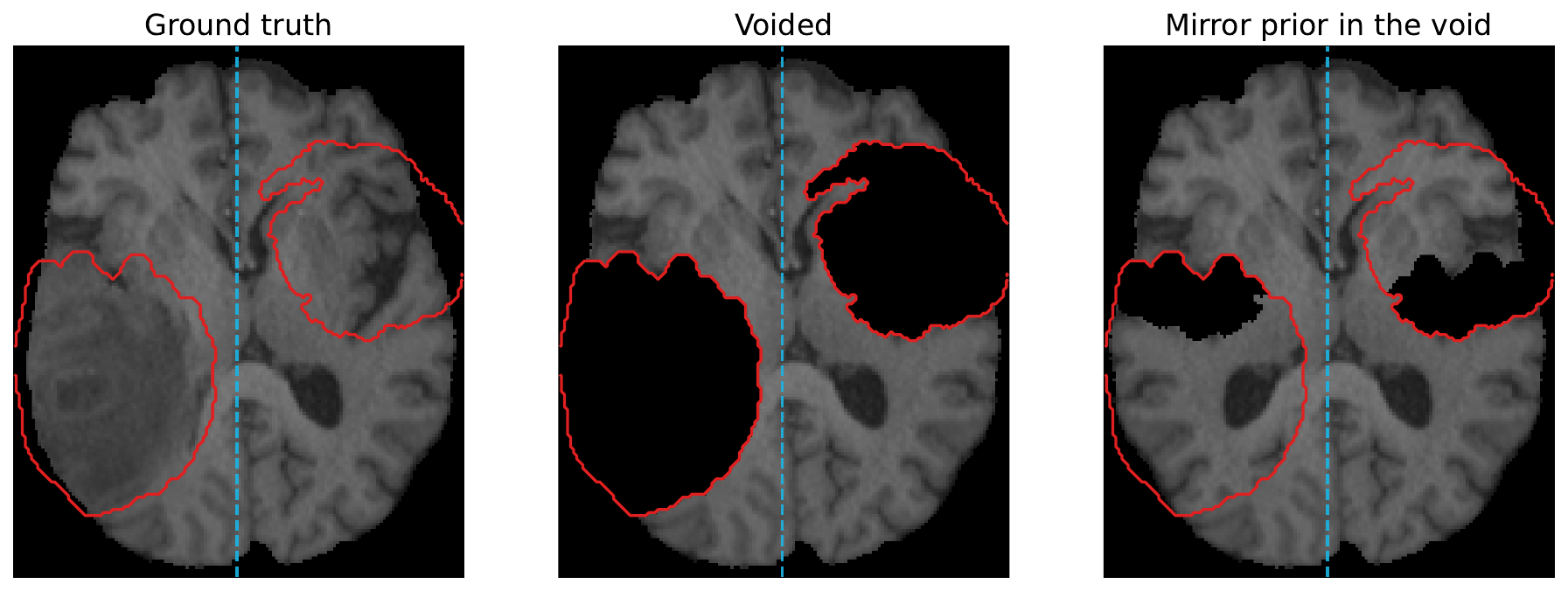}
  \caption{Contralateral-symmetry input on a double-lesion case with two voids in opposite hemispheres. \textbf{Left:} the ground-truth volume, with the two void regions outlined in red and the estimated mid-sagittal plane dashed. \textbf{Middle:} the voided input the network actually receives. \textbf{Right:} filling each void with its contralateral mirror. Because the two lesions are nearly contralateral, part of one void mirrors onto the other (or onto out-of-brain background) and is left unfilled; the validity channel (not shown) flags exactly these voxels so the network learns to ignore them, while the remaining void is given a plausible anatomical template.}
  \label{fig:contra}
\end{figure}

We condition the network on the brain's bilateral symmetry~\cite{ma2024symmetry} by supplying two additional input channels alongside the voided volume $x_v$ and mask $m$. We first estimate the mid-sagittal plane per sample as the intensity-weighted centroid along the left--right axis $h$,
\begin{equation}
  c=\frac{\sum_h h\,w_h}{\sum_h w_h}, \qquad w=(x_v+1)/2,
\end{equation}
where the weight $w$ rescales the normalised intensities to $[0,1]$ so background voxels contribute nothing. Reflecting the left-right coordinate about $c$ gives the mirror index $h'=2c-h$ (rounded to the nearest voxel), where the axis runs over $h\in\{0,\dots,H-1\}$. The mirrored volume copies the voided intensity from that index, $\tilde{x}_h = x_v[\,h'\,]$, and a per-voxel validity map verifies whether the mirror source is usable:
\begin{equation}
  \bar v_h = \begin{cases}
    1 & \text{if } 0\le h' < H \text{ and } m[\,h'\,]=0, \\
    0 & \text{otherwise,}
  \end{cases}
\end{equation}
that is, $\bar v_h=1$ only where the mirror index stays inside the volume and its source is observed tissue rather than void ($m=1$ marks the masked, to-be-synthesised voxels). Conditioning is obtained by concatenation, so the network receives the four-channel input $[\,x_v,m,\tilde{x},\bar v\,]$ and the composited completion becomes $\hat{x}=x_v\odot(1-m)+r_\theta(x_v,m,\tilde{x},\bar v)\odot m$; here $\tilde{x}$ supplies a candidate value for each masked voxel and $\bar v$ marks where that value is defined. Because homologous structures across the mid-line are nearly identical in healthy anatomy and lesions are usually unilateral~\cite{ma2024symmetry}, $\tilde{x}$ provides a template for tissue missing on one side (Fig.~\ref{fig:contra}), while $\bar v$ lets the network discount invalid (out-of-brain or itself-voided) mirror content, for example in bilateral or mid-line lesions where the mirror source is unavailable. The stem weights on the two extra channels are zero-initialised, so the model reproduces its no-contralateral counterpart and can be warm-started from it.

\subsection{Training objective and schedule}
We minimise a weighted sum of masked error terms over the void, plus a high-pass term that discourages over-smoothing: $\mathcal{L}=\lambda_1\tfrac{\sum|\hat{x}-x|\odot w}{\sum w}+\lambda_2\tfrac{\sum(\hat{x}-x)^2\odot w}{\sum w}+\lambda_s\big(1-\mathrm{SSIM}_m(\hat{x},x)\big)+\lambda_{\mathrm{hf}}\,\mathcal{L}_{\mathrm{hf}}$, with $\lambda_1{=}1$, $\lambda_2{=}10$, $\lambda_s{=}1$ and $\lambda_{\mathrm{hf}}{=}0.5$. The squared-error term dominates because the official per-case PSNR rank is exactly the inverse MSE rank, so the score is two parts MSE to one part SSIM. The two error terms use the weight map $w=m\odot\big(1-(1-\gamma)\,\mathbf{1}[\text{tumour}]\big)$ with $\gamma{=}0.25$, which down-weights the void voxels that lie on tumour tissue and are therefore never scored, whereas the SSIM term keeps the full void $m$ because it is a windowed statistic that needs the surrounding context. The high-pass term is the masked $L_1$ distance between the high-frequency residuals of prediction and target, $\mathcal{L}_{\mathrm{hf}}=\tfrac{\sum|\mathrm{HF}(\hat{x})-\mathrm{HF}(x)|\odot m}{\sum m}$ with $\mathrm{HF}(x)=x-G_{\sigma_{\mathrm{hf}}} * x$ and $G_{\sigma_{\mathrm{hf}}}$ a Gaussian of $\sigma_{\mathrm{hf}}{=}1$ voxel; unlike the other terms it is not minimised by blurring, since smoothing the prediction drives $\mathrm{HF}(\hat{x})$ towards zero and widens the gap. Augmentation follows masked image modelling~\cite{he2022mae}: instead of always re-using the provided void, at each step we draw a real mask from a bank of inpainting masks (eight per case per epoch) and crop a random window around it, so the network sees many void shapes, sizes and positions rather than memorising a single. The bank contains whole real BraTS inpainting masks, including multi-component and peri-lesional regions; masks are not split into connected components and are randomly placed within the brain. We optimise with AdamW~\cite{loshchilov2019adamw} at batch size $2$ with zero weight decay. We keep an EMA of the weights~\cite{polyak1992acceleration} (EMA decay $0.995$) for evaluation and for model selection on the mean healthy-region SSIM over $60$ held-out training cases. The training ran for up to 400 epochs with 200 steps warmup and cosine schedule peaking at $10^{-4}$ and annealed to $10^{-7}$~\cite{loshchilov2017sgdr}.

\subsection{Inference}
At test time we apply mirror test-time augmentation~\cite{wang2019tta}, averaging the prediction with its sagittal flip. Inference is a single deterministic forward pass per view (two with the flip). The composited prediction is de-normalised and written back into the original volume geometry.

\section{Experiments and Results}\label{results}

\begin{table}[t]
\centering
\caption{Top: ablation on our validation split. Middle: final architectures scored with the official metric on the native volume, mean (std) over the $60$ held-out cases. Bottom: official validation leaderboard over $219$ cases, mean (std).}
\label{tab:results}
\setlength{\tabcolsep}{5pt}
\begin{tabular}{lccc}
\toprule
Configuration & SSIM\,$\uparrow$ & PSNR\,$\uparrow$ & MSE\,$\downarrow$ \\
\midrule
Full transformer~\cite{danese2026wavedit}   & 0.580 & --    & --     \\
U-DiT backbone                 & 0.848 & 21.5  & 0.0053 \\
\; + healthy-only attention    & 0.853 & 21.9  & 0.0049 \\
\; + contralateral input       & 0.856 & 22.1  & 0.0047 \\
\; + TTA \& annealing & 0.865 & 22.5 & 0.0044 \\
\;~~~~(no contralateral attention) &&&\\
\midrule
  U-DiT (ours), official metric  & \textbf{0.880} (0.094) & \textbf{23.83} (2.58) & \textbf{0.0043} (0.0034) \\
  Wavelet U-DiT                  & 0.870 & 23.1 & 0.0048 \\
  Wavelet HDiT~\cite{danese2026wavedit} & 0.781 & 17.9 & 0.0142 \\
\midrule
  Leaderboard, $219$ cases & \textbf{0.864} (0.085) & \textbf{24.71} (4.03) & \textbf{0.0046} (0.0032) \\
\bottomrule
\end{tabular}
\end{table}

\subsection{Evaluation metrics and protocol}
The task is scored on the healthy sub-region of the mask by three distortion metrics, SSIM, PSNR and MSE, and the headline result is the score returned by the official BraTS-2026 evaluation server. For the ablation we additionally report relative comparisons on our own held-out validation split of the training data, using EMA weights and the official healthy-region convention.

\subsection{Ablation study}
Table~\ref{tab:results} isolates the effect of each design choice on the internal split, added cumulatively. Replacing the full-transformer (hierarchical hourglass with neighbourhood attention) backbone by our convolutional U-DiT hierarchy is decisive: the full transformer plateaus around $0.58$ SSIM, whereas the U-DiT backbone alone reaches $0.848$. Restricting the bottleneck attention to known-healthy tokens then improves all three metrics and is our single most effective change; adding the contralateral-symmetry input yields a further, smaller gain concentrated on the distortion metrics (PSNR, MSE) at essentially unchanged SSIM, consistent with its role as a low-frequency anatomical template; and mirror test-time augmentation together with the annealed learning-rate schedule contributes a final increment. We also evaluated two wavelet-domain variants~\cite{danese2026wavedit}: moving our backbone onto a 3D wavelet decomposition processes eight times fewer spatial positions, and replacing its bottleneck with an hourglass transformer reduces the cost further. Both are cheaper to run, but neither matches the full-resolution U-DiT on the official metric, even though the wavelet variant recovers slightly more anatomical structure inside the void (Fig.~\ref{fig:realism}), so we retain the U-DiT for the submission. We also ablated three additions that did not improve the official distortion score and were therefore excluded from the submitted model: averaging our output with a separate CNN inpainter, matching the hole mean to the surrounding tissue shell, and greedy weight-space souping~\cite{wortsman2022modelsoups} of our (strongly correlated) checkpoints. The submitted model is thus a single network with mirror test-time augmentation and contralateral attention.

\subsection{Validation leaderboard}
\begin{figure}[tb]
  \centering
  \includegraphics[width=0.7\textwidth]{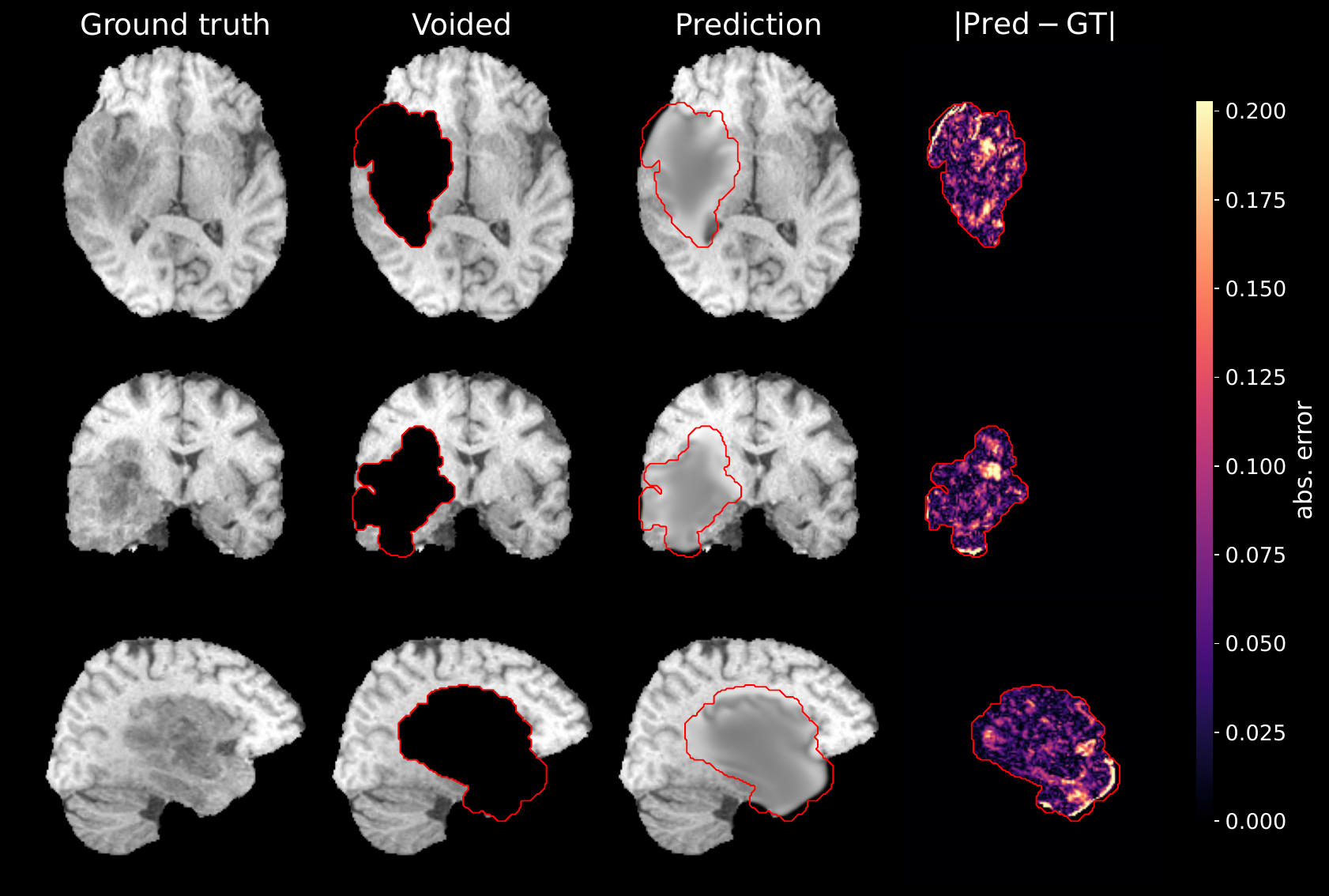}
  \caption{Qualitative completion for a validation case (our submitted mirror-TTA model). Columns, left to right: ground truth, voided input with the void boundary outlined in red, our completion and the absolute error $|\hat{x}-x|$. Rows: axial, coronal and sagittal views through the void centroid.}
  \label{fig:qual}
\end{figure}

We uploaded our predictions to the official BraTS-2026 Task-4 (Inpainting) validation queue on the challenge platform~\cite{karargyris2023medperf}. Table~\ref{tab:results} reports the scores it returns over the $219$-case validation set. The official scores differ from our internal split in both directions, with lower SSIM and higher PSNR, which is expected given the different case distributions. Figure~\ref{fig:qual} shows a representative completion: known tissue is preserved exactly, so the error is supported only inside the void.

\section{Discussion}\label{discussion}

\begin{figure}[tb]
  \centering
  \includegraphics[width=0.7\textwidth]{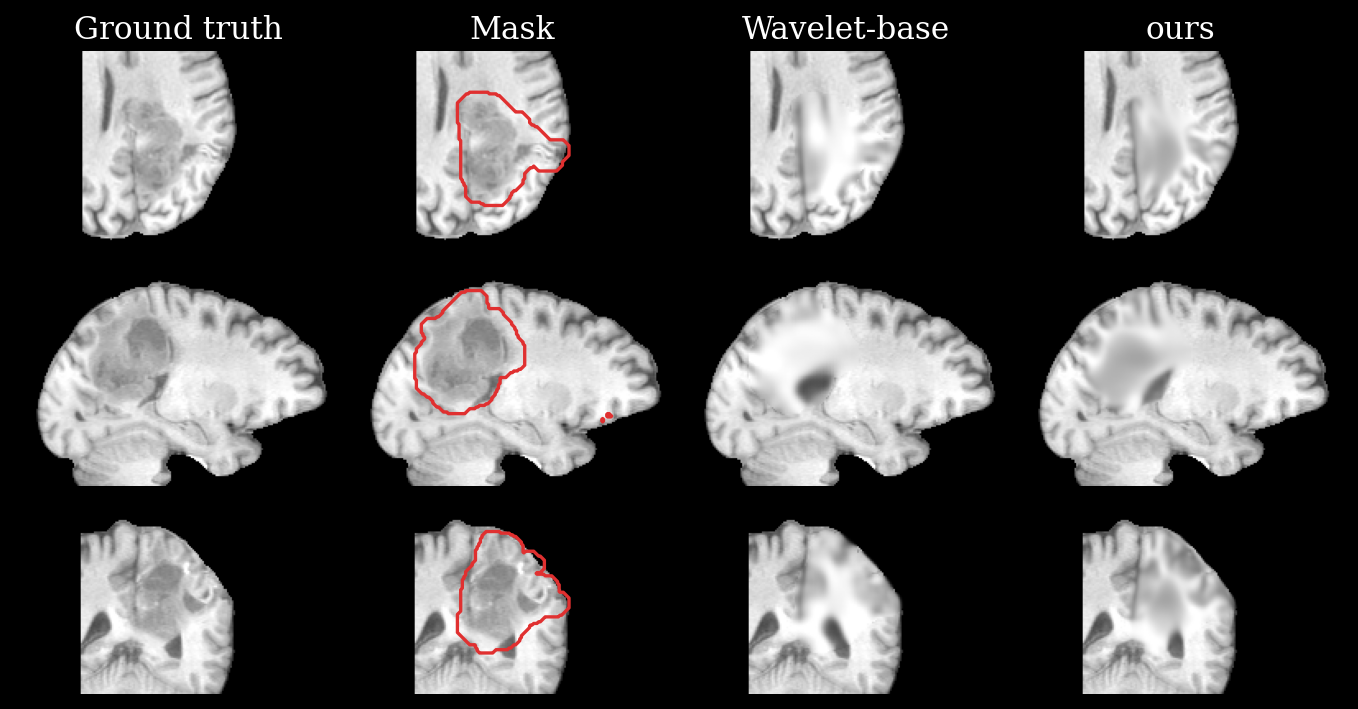}
  \caption{Perception-distortion trade-off on one case, three views. Columns, left to right: ground truth, the void mask (red), our wavelet variant and our metric-optimal completion (U-DiT). The wavelet variant recovers slightly more anatomical structure inside the void yet scores lower on all three distortion metrics, so the texture a reader judges more realistic is not what the score rewards.}
  \label{fig:realism}
\end{figure}

Our results are driven by matching the inductive bias of the network to the structure of the inpainting task. \emph{Where} global context is injected matters more than simply adding it: an unconstrained attention bottleneck moves the score little, whereas forcing occluded tokens to read only known-healthy context makes reconstruction an inference from observed anatomy rather than a fixed-point over unknown regions. This change is parameter-free and applies to any inpainting network with a coarse attention stage, which makes it a broadly reusable design. The contralateral-symmetry input is complementary: it hands the network an explicit, patient-specific template for the missing side and improves the distortion metrics at matched structural similarity. Its benefit persists alongside mirror test-time augmentation even though both exploit the brain's left-right symmetry, indicating that a symmetry prior is more effective when supplied to the model as conditioning than when applied only as a post-hoc average.

\noindent\textbf{Realism versus the distortion metric.}
Like other distortion-optimised regression models, our completions remain smooth. Since SSIM, PSNR, and MSE are optimised by the conditional mean of plausible completions, predictions tend to suppress uncertain high-frequency details while preserving low-frequency anatomical structure. Consequently, the synthesised tissue contains only a fraction of the high-frequency texture energy of real tissue: measured over the scored region with the same high-pass operator the training term uses, the completions retain $0.65\pm0.11$ of its high-frequency RMS.
This is the expected behaviour on a single-reference distortion score rather than a failure of the model, but it does mean the metric and anatomical realism can diverge (Fig.~\ref{fig:realism}). Closing this gap while preserving the distortion score, for instance with a residual generative refinement stage that adds high-frequency detail without disturbing the low-frequency content the metric rewards, is the most promising direction for future work, and motivates complementing SSIM with texture-aware measures.

\section{Conclusion}\label{conclusion}

We presented a U-DiT-style volumetric network for healthy brain-tissue inpainting, in which a single downsampled global self-attention block with 3D rotary position embeddings supplies long-range context while convolutions and skip connections preserve high-frequency detail. Two task-specific ideas drive its performance: restricting the attention so that occluded tokens read only known-healthy context, and a contralateral-symmetry input that gives the network the patient's own healthy hemisphere as a template for the missing tissue. On the official BraTS-2026 validation leaderboard the submission reaches a mean healthy-region SSIM of $0.864$, PSNR of $24.71$\,dB and MSE of $4.57{\times}10^{-3}$ over $219$ cases. Our analysis of the residual smoothness of distortion-optimal regression points to residual generative refinement as the route to greater anatomical realism without sacrificing the distortion metrics.

\subsubsection*{Acknowledgements.} We thank the BraTS challenge organisers for the data and evaluation platform. \\ Synapse team: \textbf{Morpheus}.

\bibliographystyle{splncs04}
\bibliography{ref}

\end{document}